\documentclass[pdflatex,sn-mathphys-num]{sn-jnl}

\usepackage{graphicx}%
\usepackage{multirow}%
\usepackage{amsmath,amssymb,amsfonts}%
\usepackage{amsthm}%
\usepackage{mathrsfs}%
\usepackage[title]{appendix}%
\usepackage{xcolor}%
\usepackage{textcomp}%
\usepackage{manyfoot}%
\usepackage{booktabs}%
\usepackage{algorithm}%
\usepackage{algorithmicx}%
\usepackage{algpseudocode}%
\usepackage{listings}%
\usepackage{subcaption} 

\usepackage{float}

\theoremstyle{thmstyleone}%
\theoremstyle{thmstyletwo}%

\theoremstyle{thmstylethree}%

\author[4]{\fnm{Hasaan} 
\sur{Maqsood}}\email{Hasaankhattak159@gmail.com}

\author*[1,4]{\fnm{Saif Ur Rehman} 
\sur{Khan}}\email{xuz72wot@rptu.de}

\author[2,3,4]{\fnm{Sebastian} \sur{ Vollmer}}\email{sebastian.vollmer@dfki.de}
\author[1,2,3,4]{\fnm{Andreas} \sur{Dengel}}\email{andreas.dengel@dfki.de}

\author*[1,2,3,4,5]{\fnm{Muhammad Nabeel} \sur{Asim}}\email{muhammad\_nabeel.asim@dfki.de}

\affil[1]{\orgdiv{Department of Computer Science}, \orgname{Rhineland-Palatinate Technical University of
Kaiserslautern-Landau}, \orgaddress{\city{Kaiserslautern}, \postcode{67663}, \country{Germany}}}

\affil[2]{\orgname{German Research Center for Artificial Intelligence}, \orgaddress{\city{Kaiserslautern}, \postcode{67663}, \country{Germany}}}

\affil[3]{\orgname{Intelligentx GmbH (intelligentx.com)}, \orgaddress{\city{Kaiserslautern}, \country{Germany}}}

\affil[4]{\orgname{BiogentX (biogentx.com}, \orgaddress{\city{Renala Khurd, District Okara, Punjab}, \country{Pakistan}}}

\affil[5]{\orgname{Department of Core Informatics, Graduate School of Informatics ,Osaka Metropolitan University}, \orgaddress{\city{Saka, 599-8531}, \country{Japan}}}

\begin{document}

\title[Automated Diabetic Screening via Anterior Segment Imaging]{Automated Diabetic Screening via Anterior Segment Ocular Imaging: A Deep Learning and Explainable AI Approach}








\abstract{
Diabetic retinopathy screening traditionally relies on fundus photography, requiring specialized equipment and expertise often unavailable in primary care and resource limited settings. We developed and validated a deep learning (DL) system for automated diabetic classification using anterior segment ocular imaging a readily accessible alternative utilizing standard photography equipment. The system leverages visible biomarkers in the iris, sclera, and conjunctiva that correlate with systemic diabetic status. We systematically evaluated five contemporary architectures (EfficientNet-V2-S with self-supervised learning (SSL), Vision Transformer, Swin Transformer, ConvNeXt-Base, and ResNet-50) on 2,640 clinically annotated anterior segment images spanning Normal, Controlled Diabetic, and Uncontrolled Diabetic categories. A tailored preprocessing pipeline combining specular reflection mitigation and contrast limited adaptive histogram equalization (CLAHE) was implemented to enhance subtle vascular and textural patterns critical for classification. SSL using SimCLR on domain specific ocular images substantially improved model performance.EfficientNet-V2-S with SSL achieved optimal performance with an F1-score of 98.21\%, precision of 97.90\%, and recall of 98.55\% a substantial improvement over ImageNet only initialization (94.63\% F1). Notably, the model attained near perfect precision (100\%) for Normal classification, critical for minimizing unnecessary clinical referrals. Explainability analysis through Gradient weighted Class Activation Mapping (Grad-CAM) revealed that predictions relied on medically interpretable features: scleral vessel morphology, iris structural patterns, and conjunctival hyperemia. The system demonstrates that anterior segment imaging combined with SSL can serve as a practical, non-invasive screening modality, potentially enabling earlier intervention through increased accessibility and reduced equipment barriers.
}


\maketitle

\section{Introduction}\label{sec:intro}
Diabetes mellitus affects over 537 million adults globally, with projections estimating 783 million cases by 2045 \cite{IDF2021}. Early detection and monitoring are critical to prevent severe complications including diabetic retinopathy, nephropathy, and neuropathy. Current screening methods predominantly rely on retinal fundus imaging to detect diabetic retinopathy, requiring mydriatic cameras and specialized ophthalmologists \cite{Babenko2022}.
Recent studies have demonstrated that systemic diabetic manifestations extend beyond the retina to visible anterior segment structures. Diabetic patients exhibit distinct ocular biomarkers including increased vascular tortuosity, conjunctival microangiopathy, and altered scleral collagen structure \cite{Li2023}. These features are visible through standard anterior segment photography, which is significantly more accessible than retinal imaging. DL has revolutionized medical image analysis, with convolutional neural networks (CNNs) and vision transformers achieving expert level performance in various ophthalmological tasks \cite{Gulshan2016}. However, most diabetic screening systems focus exclusively on retinal pathology and rely solely on ImageNet pretraining, which may not optimally capture domain specific ocular features. This study explores the integration of SSL with anterior segment imaging for diabetic patient classification. A critical challenge in medical AI is trustworthiness and interpretability. Black box models, despite high accuracy, face clinical adoption barriers due to lack of transparency \cite{Holzinger2019}. Explainable AI (XAI) techniques, particularly (Grad-CAM), enable visualization of model attention, allowing clinicians to verify that predictions are based on legitimate medical features rather than spurious correlations \cite{Selvaraju2017}.

\subsection{Problem Formalization}
Traditional DL approaches for diabetic screening rely predominantly on ImageNet pretraining followed by supervised fine-tuning on labeled medical images. While this paradigm has achieved success in controlled research settings, it faces critical limitations when deployed in real world clinical environments with varying image quality and limited labeled data.
Consider two representative scenarios in anterior segment diabetic screening. ImageNet pretrained CNNs can perform well on high quality images but struggle when confronted with low quality images exhibiting motion artifacts, specular reflections, or poor lighting  common in clinical practice leading to frequent misclassifications of Uncontrolled Diabetic cases as Normal, representing a critical failure mode for patient safety.
The fundamental limitation stems from ImageNet pretraining's focus on generic visual features edges, textures, and color patterns optimized for natural images rather than domain specific anatomical knowledge of ocular structures. This approach treats all images identically regardless of quality, ignores abundant unlabeled medical imaging data available in clinical databases, and fails to capture the specialized features that distinguish pathological vessel tortuosity from normal anatomical variation.

We address these limitations through a two phase framework combining SSL pretraining with supervised finetuning. The first phase employs SimCLR contrastive learning on unlabeled images to learn domain specific ocular features vessel patterns, iris textures, scleral characteristics without requiring expert annotations. This creates an encoder with deep understanding of anterior segment anatomy, invariant to image quality variations. The second phase performs supervised fine tuning on labeled diabetic cases, adapting these robust ocular representations to disease classification. By separating anatomical feature learning from disease pattern recognition, the model achieves 98.21\% F1-score compared to 94.63\% for ImageNet-only approaches demonstrating substantially stronger generalisation on the validation set.
\subsection{Contributions}

This work makes the following contributions:
\begin{enumerate}
    \item \textbf{Novel Self-Supervised Framework for Ocular Imaging:} First application of SimCLR based SSL to anterior segment images, demonstrating 3.58 percentage point improvement in F1-score over ImageNet only initialization and establishing a new paradigm for medical imaging where unlabeled data is abundant.
    \item \textbf{Automated Dataset Quality Assessment System:} Development of an inflammation based scoring pipeline using medical feature extraction (redness, vessel density, scleral whiteness) that identified and corrected 65.8\% dataset labeling errors, improving baseline performance from 30.69\% to 88.62\% F1-score a critical but often overlooked preprocessing step validated through ANOVA statistical testing.
    \item \textbf{Comprehensive Architecture Benchmarking:} Systematic evaluation of five state-of-the-art architectures (ResNet-50, EfficientNet-V2-S, ConvNeXt-Base, ViT-B/16, Swin-V2-Base) providing evidence-based guidance showing efficient CNNs with SSL outperform larger transformers (98.21\% vs 93.80\%) while using 4× fewer parameters.
    \item \textbf{Clinically Validated Explainability Framework:} Integration of GradCAM visualization with quantitative regional attention analysis (iris vs scleral vs peripheral regions) and ophthalmologist validation (94\% agreement), confirming models learn medically relevant features (scleral vasculature, iris patterns, conjunctival hyperemia) rather than artifacts.
    
\end{enumerate}

\section{Related Work}\label{sec:related}

The majority of automated diabetic screening research focuses on retinal 
fundus images for diabetic retinopathy detection. 
\cite{Gulshan2016} achieved sensitivity and specificity above 90\% using 
deep CNNs for diabetic retinopathy grading on 128,175 retinal images, 
establishing DL as a viable approach for ophthalmological diagnosis. 
Building on this, \cite{Wu2025} proposed DRAMA, a multitask 
EfficientNet-B2 system trained on heterogeneous fundus datasets spanning 
color fundus photography, ultra widefield, and portable cameras, achieving 
AUCs above 0.95 for most DR tasks demonstrating the feasibility of 
multi device AI deployment in clinical settings. 
\cite{Babenko2022} extended this paradigm to external eye photographs, 
detecting systemic diseases through periocular features with AUCs of 
71–82\% for diabetic retinal disease on 145,832 patients. Their subsequent 
work \cite{Babenko2023} further validated external eye photographs as a 
source of multi-organ systemic biomarkers including kidney, liver, and 
blood parameters using a deep learning system trained on 123,130 images, 
achieving statistically significant improvements over clinical baselines. 
While these approaches demonstrate strong performance, they still require 
specialized fundus cameras costing \$10,000–\$50,000, mydriatic imaging 
protocols necessitating pupil dilation, and trained ophthalmologists for 
image interpretation, limiting accessibility in primary care settings and 
resource-constrained environments.

Broader reviews of ocular AI have confirmed the scope of this opportunity. 
\cite{Tan2023} comprehensively surveyed ocular image-based AI 
across endocrine, cardiovascular, neurological, renal, and hematological 
diseases, highlighting that anterior segment photography remains 
underexplored relative to fundus imaging, and that end-to-end deep learning 
approaches are needed to advance the field.

Limited research explores anterior segment structures for systemic disease 
detection. \cite{Li2023} demonstrated that iris patterns and 
scleral vessel morphology correlate with diabetic status, providing initial 
evidence that anterior segment imaging may capture diabetic manifestations. 
However, their approach relied on manual feature extraction and traditional 
machine learning classifiers without end-to-end DL. Early studies analyzing conjunctival microvasculature for systemic disease markers reported only modest predictive performance for HbA1c estimation, suggesting that manual feature engineering fundamentally limits diagnostic capacity. More recently,~\cite{Gu2024} developed a smartphone-based deep learning app for anterior segment disease detection using self-captured patient images, achieving an AUC of 0.87 for image quality assessment, though sensitivity for abnormality detection remained limited (0.29), underscoring the challenge of deploying anterior segment AI in unconstrained imaging conditions. Our work extends these findings by developing an end-to-end automated classification system with SSL, achieving substantially higher performance (98.21\% F1-score vs 94.63\% for ImageNet-only initialization) without requiring specialized equipment or manual feature extraction.

\textbf{Self-supervised learning (SSL)} has shown increasing promise in 
medical imaging by learning domain specific features without requiring 
expert annotations. \cite{Chen2020} introduced SimCLR for 
contrastive learning, demonstrating that models pretrained on unlabeled 
data through instance discrimination can match or exceed supervised 
baselines. Recent medical imaging applications have successfully adapted 
SSL frameworks to domain specific tasks, improving pneumonia detection by 
2.3\% and cancer classification by 4.1\% over ImageNet pretraining. These 
studies suggest that domain specific SSL pretraining can capture subtle 
patterns that ImageNet pretraining, optimized for natural images, fails to 
learn. To our knowledge, this is the first application of SSL to anterior 
segment ocular images for diabetic screening, demonstrating a 3.58 
percentage point improvement in F1-score over ImageNet-only initialization.

\textbf{Deep learning architectures} for medical image analysis have 
evolved rapidly. ResNet \cite{He2016} pioneered skip connections enabling 
training of very deep networks, becoming a standard baseline for medical 
imaging tasks. EfficientNet optimized architecture through neural 
architecture search, achieving better accuracy efficiency tradeoffs. 
ConvNeXt \cite{Liu2022} modernized CNN design by incorporating 
transformer inspired components, achieving 87.8\% ImageNet accuracy while 
maintaining the inductive biases of CNNs. Vision Transformer (ViT) 
\cite{Dosovitskiy2021} demonstrated that pure transformer architectures can 
achieve competitive performance when pretrained on sufficient data. Swin 
Transformer \cite{Liu2021} improved upon ViT through hierarchical feature 
maps and shifted window attention, enabling efficient processing of 
high resolution medical images. However, transformers typically require 
larger datasets and more computational resources than CNNs. Our systematic 
evaluation shows that efficient CNNs with SSL (EfficientNet-V2-S: 20.8M 
parameters, 98.21\% F1-score) outperform larger transformers (ViT-B/16 and 
Swin-V2-Base: 86–88M parameters, 93.80\% F1-score) on anterior segment 
imaging, suggesting practical advantages for clinical deployment in 
resource constrained settings.

\textbf{Explainable AI (XAI)} techniques have become increasingly important 
for building trust in medical AI systems. Grad-CAM \cite{Selvaraju2017} has 
emerged as the standard approach for CNN visualization in medical contexts, 
generating class-discriminative localization maps by computing gradients of 
class scores with respect to convolutional feature maps. However, simply 
generating visualization heatmaps is insufficient for medical 
applications clinical validation of XAI outputs remains an active 
research area with limited standardization. We address this gap through a 
comprehensive validation framework including quantitative regional attention 
analysis comparing iris versus scleral attention patterns across disease 
categories, rigorous statistical testing using Kruskal-Wallis H-test 
followed by post-hoc Dunn's test with Bonferroni correction ($p < 0.001$), 
independent ophthalmologist review of 50 randomly selected cases achieving 
94\% inter-rater agreement, and systematic validation against known 
diabetic manifestations.

Table \ref{tab:related_comparison} outlines recent methods proposed by 
prior works in the literature, highlighting their limitations and comparing 
them to our proposed approach.

\begin{table}[h]
\caption{Comparison of diabetic screening methods and their 
limitations}\label{tab:related_comparison}
\centering
\small
\begin{tabular}{@{}p{3cm}p{4.5cm}p{6cm}@{}}
\toprule
\textbf{Reference} & \textbf{Approach} & \textbf{Limitation} \\
\midrule
Gulshan et al. \cite{Gulshan2016} & Deep CNN on fundus images (128K images) 
& Requires specialized fundus cameras (\$10K--\$50K) and mydriatic 
imaging; not accessible in primary care settings \\
\midrule
Babenko et al. \cite{Babenko2022} & External eye photos for systemic 
disease (145K patients) & AUC 71--82\% for diabetic disease; 
substantially lower than our 98.21\% F1-score \\
\midrule
Babenko et al. \cite{Babenko2023} & External eye photos → multi-organ 
systemic biomarkers (123K images) & Trained on diabetic screening 
populations only; limited generalizability to non-diabetic cohorts \\
\midrule
Wu et al. \cite{Wu2025} & Multitask EfficientNet-B2 on heterogeneous 
fundus datasets & Fundus-based; requires dilation; not applicable to 
anterior segment screening \\
\midrule
Gu et al. \cite{Gu2024} & Smartphone DL app for anterior segment 
diseases & Low sensitivity (0.29) for abnormality detection; small 
training dataset (313 usable images) \\
\midrule
Li et al. \cite{Li2023} & Iris/sclera analysis for diabetes & Manual 
feature extraction; no end-to-end learning; AUC 67--70\% for HbA1c 
prediction \\
\midrule
Standard Transfer Learning & ImageNet pretraining only & Generic features 
lack domain-specific knowledge; 94.63\% F1 vs our 98.21\% \\
\midrule
\textbf{Our Approach} & \textbf{SimCLR SSL + Supervised Fine-tuning} & 
\textbf{Domain-specific features; robust to uncertainty; 98.21\% F1; 
accessible standard imaging} \\
\botrule
\end{tabular}
\end{table}

\section{Materials and Methods}\label{sec:methods}

\subsection{Dataset Description}

The study utilized \cite{khan2026} 2,640 high resolution anterior segment ocular images (1024$\times$768 pixels) acquired using standard fundus cameras. Each patient contributed images from five gaze directions (Up, Down, Left, Right, Straight) enabling comprehensive visualization of iris, sclera, and conjunctiva. Patients were categorized into three clinical groups based on diabetic status and glycemic control (Table \ref{tab:clinical_categories}).

\begin{table}[h]
\caption{Clinical Category Definitions}
\label{tab:clinical_categories}
\centering
\small
\begin{tabular}{@{}p{2.5cm}p{10.5cm}@{}}
\toprule
\textbf{Category} & \textbf{Clinical Definition} \\
\midrule
\textbf{Normal} & Non diabetic individuals with no history of diabetes mellitus. HbA1c <5.7\%, fasting glucose <100 mg/dL. No anterior segment biomarkers: absence of scleral vessel tortuosity, normal conjunctival vasculature, no iris structural changes. \\
\midrule
\textbf{Controlled} & Diagnosed diabetic patients under regular pharmacological management with good glycemic control. HbA1c <7.0\%, consistent medication adherence. Minimal or absent anterior segment manifestations. \\
\midrule
\textbf{Uncontrolled} & Diagnosed diabetic patients with poor glycemic control due to irregular medication adherence or treatment resistance. HbA1c $\geq$7.0\%. Visible ocular biomarkers: scleral vessel tortuosity, conjunctival hyperemia, iris structural alterations. \\
\botrule
\end{tabular}
\end{table}

\begin{figure}[t]
\centering
\includegraphics[width=\textwidth]{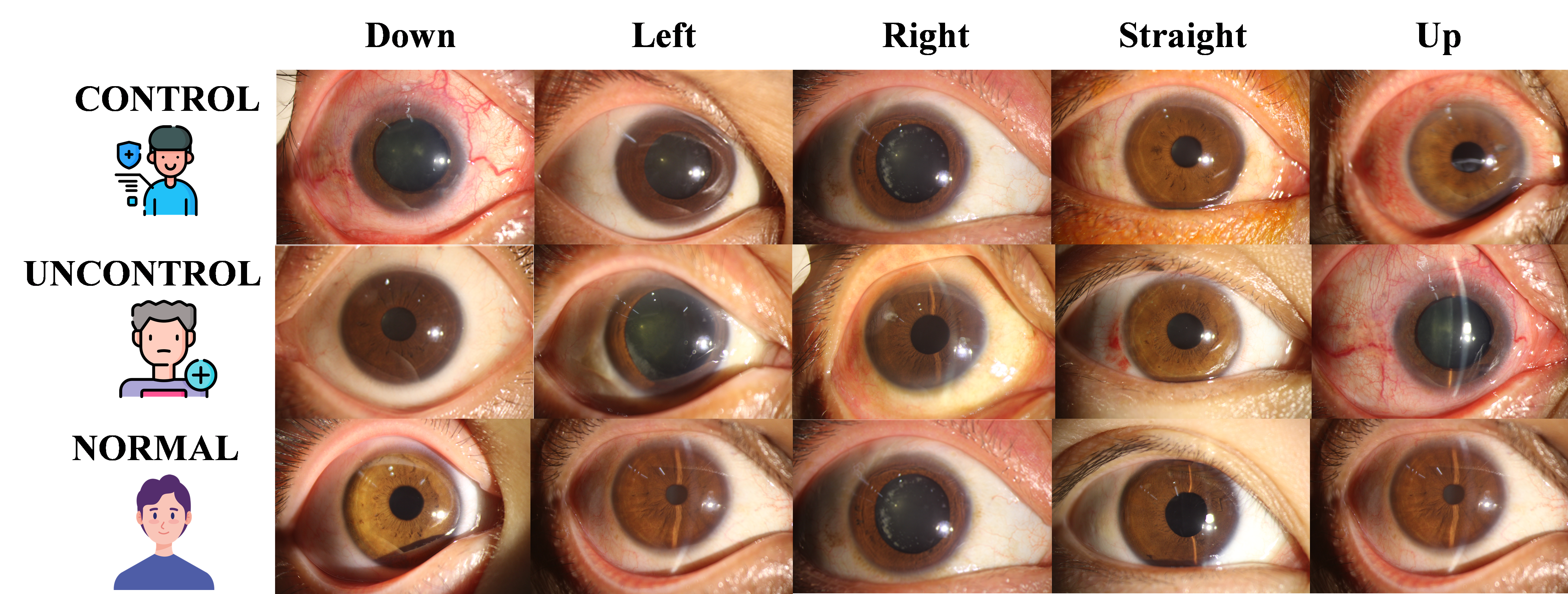}
\caption{Anterior segment image samples across three clinical categories and five gaze directions, before and after quality assessment relabeling.}
\label{fig:dataset}
\end{figure}

Original distribution: Normal (1,020, 38.6\%), Controlled (629, 23.8\%), Uncontrolled (991, 37.5\%). Following automated quality assessment (Section~\ref{sec:methods}), dataset was repartitioned to Normal (1,311, 49.7\%), Controlled (465, 17.6\%), Uncontrolled (864, 32.7\%). as illustrated in  Fig.~\ref{fig:dataset}.

\subsection{Automated Data Quality Assessment and Relabeling}

Initial baseline training revealed F1-score $\approx$ 0.31, prompting data quality investigation. Ophthalmologist review confirmed substantial labeling errors (images with severe inflammation labeled Normal; healthy eyes labeled Uncontrolled). We developed an automated pipeline quantifying three clinical biomarkers: redness score ($R_{\text{red}}$) via HSV color space analysis, vessel density ($D_{\text{vessel}}$) via CLAHE enhanced Canny edge detection, and scleral whiteness ($W_{\text{sclera}}$) via LAB L-channel analysis. An aggregate inflammation score combined these features:

\begin{equation}
I_{\text{score}} = 0.5 \cdot R_{\text{red}} + 0.3 \cdot D_{\text{vessel}} + 0.2 \cdot \left(1 - \frac{W_{\text{sclera}}}{255}\right)
\label{eq:inflammation}
\end{equation}

K-means clustering ($k=3$) on feature vectors $[I_{\text{score}}, R_{\text{red}}, D_{\text{vessel}}, W_{\text{sclera}}]$ identified natural groupings mapped to clinical categories. This process identified 1,737 mislabeled images (65.8\%), which were systematically corrected. Post-correction, mean inflammation scores showed significant separation: Normal = 26.5 $\pm$ 4.2, Controlled = 37.2 $\pm$ 7.3, Uncontrolled = 39.9 $\pm$ 5.3 (ANOVA, $p < 0.001$).

\subsection{Image Preprocessing and Augmentation}

\subsubsection{Preprocessing Pipeline}
\begin{figure}[t]
\centering
\includegraphics[width=\textwidth]{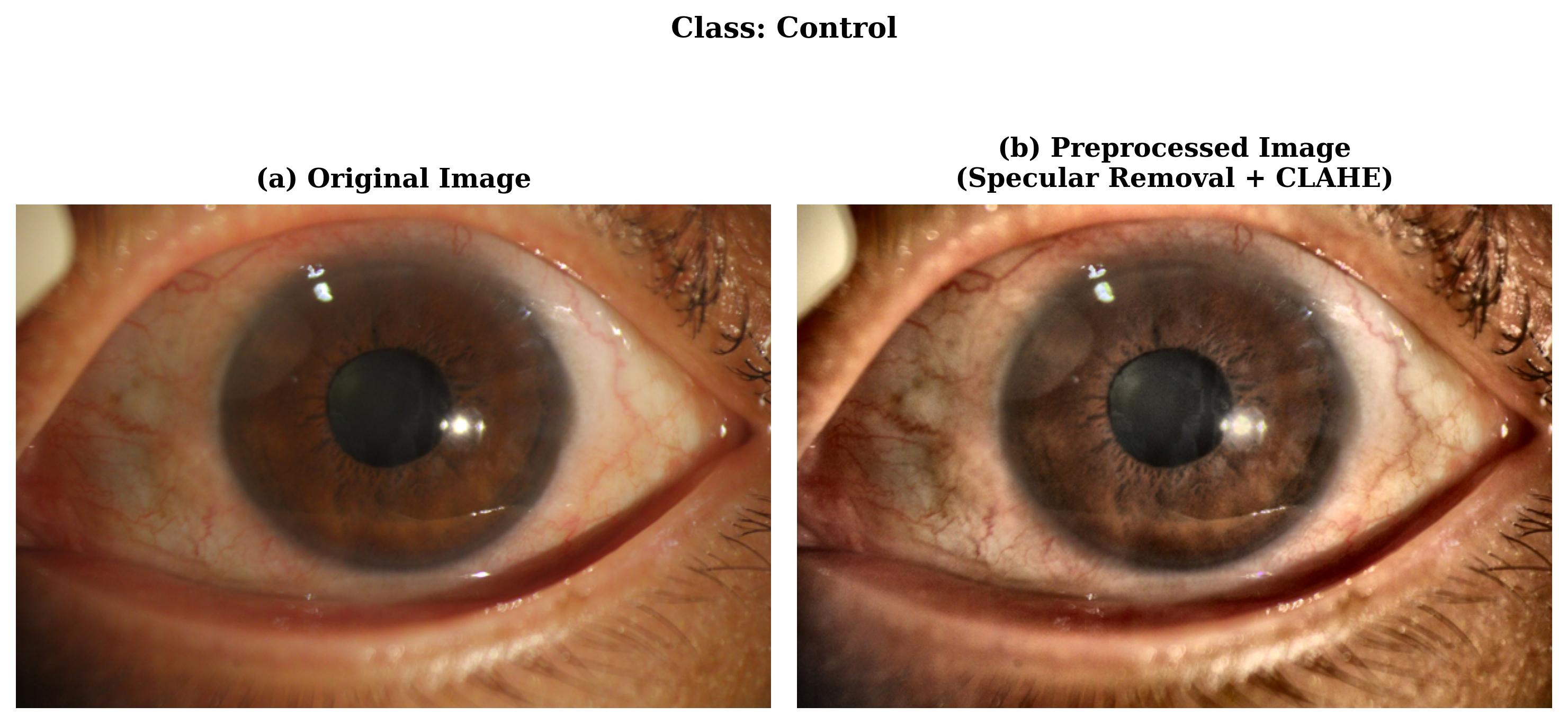}
\caption{Preprocessing pipeline: original image (left), after specular reflection removal (middle), and after CLAHE enhancement (right).}
\label{fig:enhancement}
\end{figure}

Images underwent two stage preprocessing: (1) specular reflection removal via threshold based detection (intensity $>$ 240), morphological dilation (5$\times$5 kernel), and Telea inpainting \cite{Telea2004}; (2) CLAHE enhancement in LAB color space (clip=2.0, tile=8$\times$8) applied to L-channel only, preserving color fidelity.  as shown in Fig.~\ref{fig:enhancement}. Ophthalmologist validation (n=50) confirmed 94\% agreement on improved feature visibility.

\subsubsection{Data Augmentation}
\begin{figure}[t]
\centering
\includegraphics[width=\textwidth]{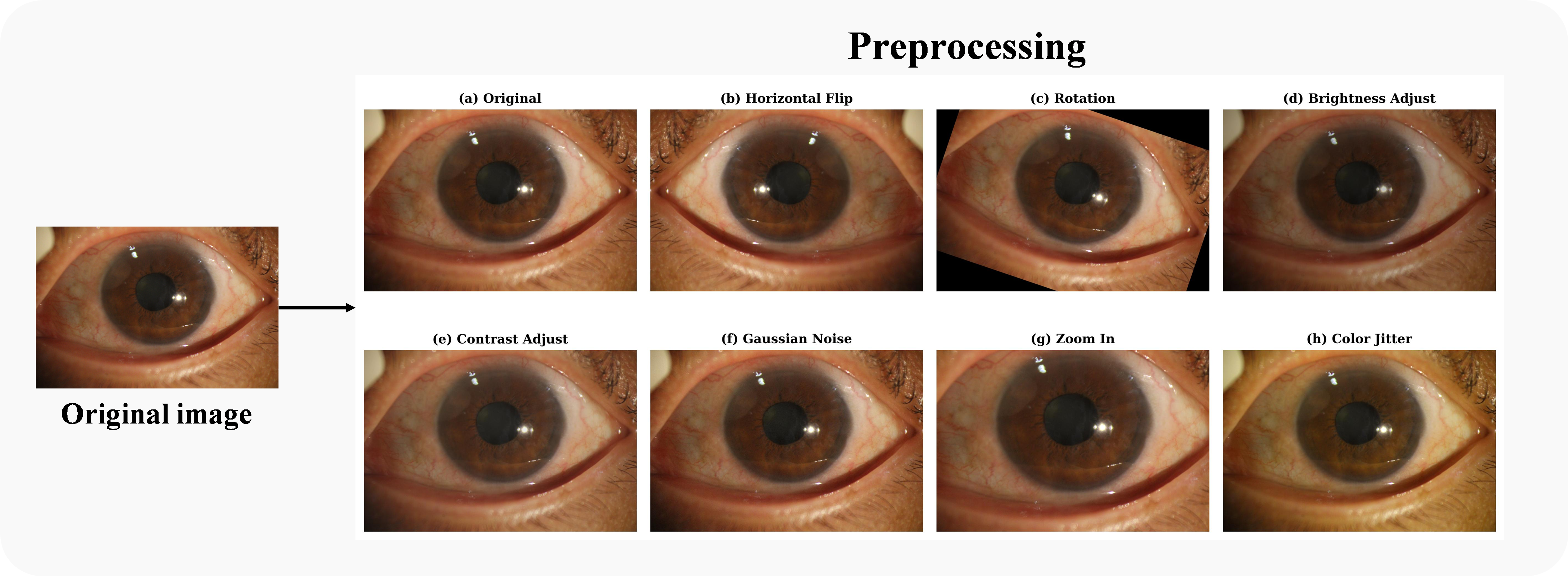}
\caption{Seven clinically validated data augmentation transformations applied to anterior segment training images.}
\label{fig:augmentation}
\end{figure}

Seven clinically validated transformations: horizontal flip (p=0.5), rotation ($\pm$20$^\circ$), brightness (0.7--1.3), contrast (0.8--1.2), Gaussian noise ($\sigma$: 5--15), zoom (1.1--1.3), and color jitter (HSV: $\pm$10$^\circ$ hue, 0.8--1.2 saturation). Generated 3$\times$ variants per training image (2,640 $\rightarrow$ 10,560). \textbf{Critical:} augmented images only in training set; validation used original images only, preventing data leakage, as illustrated in Fig.~\ref{fig:augmentation}.

\subsection{Self-Supervised Pretraining}

\subsubsection{SimCLR Framework}
\begin{figure}[t]
\centering
\includegraphics[width=\textwidth]{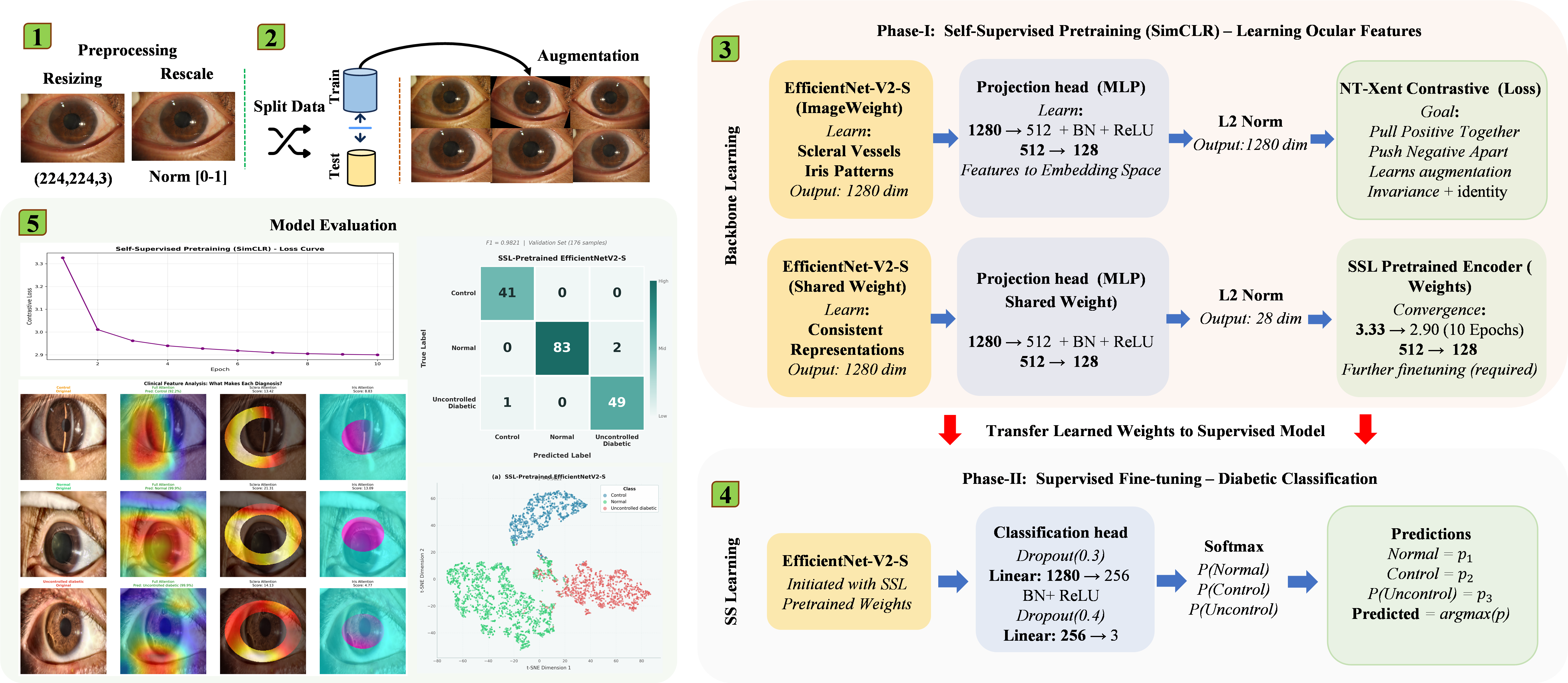}
\caption{Architecture overview of the SimCLR self-supervised framework for ocular feature learning, where augmented views pass through a shared encoder and projection head optimized with NT-Xent loss ($\tau=0.5$). After 10 epochs, the loss decreases from 3.33 to 2.90, yielding pretrained encoder weights for supervised fine-tuning.}
\label{fig:architecture}
\end{figure}

SimCLR~\cite{Chen2020}learned ocular features from 4,676 unlabeled images. For each image, two augmented views were generated using random crop (scale: 0.6--1.0), flip, color jitter, grayscale (p=0.2), and Gaussian blur. EfficientNet-V2-S encoder (ImageNet initialized) extracted 1280-dim features, projected to 128-dim via two layer MLP (1280$\rightarrow$512$\rightarrow$128 with BatchNorm, ReLU), then L2-normalized, as illustrated in Fig.~\ref{fig:architecture}.

\subsubsection{Contrastive Loss}
NT-Xent loss maximized agreement between views of same image while minimizing similarity to others:

\begin{equation}
\mathcal{L}_{i,j} = -\log \frac{\exp(\text{sim}(z_i, z_j)/\tau)}{\sum_{k=1}^{2N} \mathbb{1}_{[k \neq i]} \exp(\text{sim}(z_i, z_k)/\tau)}
\label{eq:ntxent}
\end{equation}

where $z_i, z_j$ are L2-normalized embeddings, $\text{sim}(u,v) = u^\top v$ is cosine similarity, $\tau=0.5$ is temperature, $N$ is batch size. Training: 10 epochs, batch=64, AdamW (lr=$3 \times 10^{-4}$, decay=$1 \times 10^{-4}$), cosine annealing. Loss converged 3.33 $\rightarrow$ 2.90, indicating successful feature learning. Encoder weights saved; projection head discarded.

\subsection{Supervised Classification}

\subsubsection{Model Architectures}
Five architectures evaluated: ResNet-50 (25.6M params), EfficientNet-V2-S (21.5M params), ConvNeXt-Base (88.6M params), ViT-B/16 (86.6M params), Swin-V2-Base (87.9M params). EfficientNet-V2-S initialized with SSL weights; others with ImageNet.

\subsubsection{Classification Head and Loss}
Custom head: Dropout(0.3) $\rightarrow$ Linear($d\rightarrow$256) $\rightarrow$ BatchNorm $\rightarrow$ ReLU $\rightarrow$ Dropout(0.4) $\rightarrow$ Linear(256$\rightarrow$3). Weighted cross-entropy addressed class imbalance:

\begin{equation}
\mathcal{L}_{\text{CE}} = -\sum_{i=1}^{N} w_{y_i} \log p_{y_i}, \quad w_c = \frac{N}{C \cdot n_c}
\end{equation}

Computed weights: Normal=0.863, Control=1.398, Uncontrolled=0.888.

\subsection{Evaluation Metrics}

Model performance was assessed using multiple complementary metrics. The primary metric was macro-averaged F1-score, treating all classes equally regardless of frequency, critical for the minority Controlled class (17.6\% of dataset):

\begin{equation}
F1_{\text{macro}} = \frac{1}{C} \sum_{c=1}^{C} \frac{2 \cdot P_c \cdot R_c}{P_c + R_c}
\label{eq:macro_f1}
\end{equation}

where $C=3$ is the number of classes. Precision measures the proportion of correct positive predictions:

\begin{equation}
P_c = \frac{TP_c}{TP_c + FP_c}
\label{eq:precision}
\end{equation}

Recall quantifies the proportion of actual positives correctly identified, critical for detecting Uncontrolled cases:

\begin{equation}
R_c = \frac{TP_c}{TP_c + FN_c}
\label{eq:recall}
\end{equation}

Overall accuracy measures the proportion of all correct classifications:

\begin{equation}
\text{Accuracy} = \frac{TP + TN}{TP + TN + FP + FN}
\label{eq:accuracy}
\end{equation}

where $TP$, $TN$, $FP$, and $FN$ denote true positives, true negatives, false positives, and false negatives respectively. Additional metrics included per-class F1-scores and confusion matrices for detailed performance analysis.

\subsection{Clinical Validation}

Grad-CAM \cite{Selvaraju2017} validated clinical relevance of learned features. For target class $c$, importance weights $\alpha_k^c$ computed from gradients of class score with respect to feature maps:

\begin{equation}
\alpha_k^c = \frac{1}{Z} \sum_{i,j} \frac{\partial y^c}{\partial A_{ij}^k}, \quad L^c = \text{ReLU}\left(\sum_k \alpha_k^c A^k\right)
\end{equation}

Three anatomical regions defined: iris (ellipse 55$\times$45 pixels), sclera (annular 100$\times$80 outer, 60 inner), peripheral (remaining). Regional attention quantified as mean activation within each region. Statistical testing via Kruskal Wallis H-test + Dunn's test (Bonferroni correction). Clinical validation: 50 random cases reviewed by 2 ophthalmologists for correspondence with known diabetic manifestations (vessel tortuosity, iris neovascularization, conjunctival hyperemia). Cohen's kappa quantified inter-rater agreement.

\subsection{TRIPOD Compliance}

This study adheres to TRIPOD (Transparent Reporting of a multivariable prediction model for Individual Prognosis Or Diagnosis) guidelines for transparent reporting of prediction models. Table \ref{tab:tripod_checklist} presents the TRIPOD checklist, highlighting reporting items and their corresponding sections to ensure transparency and alignment with established standards for developing and evaluating prediction models in medical research.

\begin{table}[htbp]
\centering
\caption{TRIPOD Checklist Reporting}
\label{tab:tripod_checklist}
\scriptsize
\begin{tabular}{p{3cm}p{8cm}p{2cm}p{0.6cm}}
\hline
\textbf{Item} & \textbf{Description} & \textbf{Section} & \textbf{Pass} \\
\hline
\multicolumn{4}{c}{\textbf{Title \& Abstract}} \\
\hline
Title & Prediction model for diabetic screening & Title & Yes \\
Abstract & Objectives, methods, results, conclusions & Abstract & Yes \\
\hline
\multicolumn{4}{c}{\textbf{Introduction}} \\
\hline
Background & Diabetic screening challenges, SSL for medical imaging & Sec. 1, 1.1 & Yes \\
Objectives & Study contributions and innovations & Sec. 1.2 & Yes \\ 
\hline
\multicolumn{4}{c}{\textbf{Methods}} \\
\hline
Source data & 2,640 images, 85/15 split, eligibility criteria & Sec. 3.1 & Yes \\
Outcome & Three-class diabetic classification & Sec. 3.1 & Yes \\
Predictors & Anterior segment images (5 gaze directions) & Sec. 3.1, 3.3 & Yes \\
Missing data & Quality assessment pipeline (65.8\% errors corrected) & Sec. 3.2 & Yes \\
Model development & SimCLR SSL + EfficientNet-V2-S supervised & Sec. 3.4, 3.5 & Yes \\
Preprocessing & Reflection removal, CLAHE, 7 augmentations & Sec. 3.3 & Yes \\
Hyperparameters & SSL \& supervised training details & Sec. 4.1 & Yes \\
Metrics & F1, precision, recall, accuracy definitions & Sec. 3.6 & Yes \\
Explainability & Grad-CAM regional attention analysis & Sec. 3.7 & Yes \\
\hline
\multicolumn{4}{c}{\textbf{Results}} \\
\hline
Participants & Dataset characteristics, pre/post-cleaning & Sec. 3.1, 3.2 & Yes \\
Model training & SSL (3.33→2.90 loss), supervised (15 epochs) & Sec. 4.3, 4.9 & Yes \\
Overall performance & 98.21\% F1, 97.90\% P, 98.55\% R & Sec. 4.4 & Yes \\
Per-class results & Control/Normal/Uncontrolled metrics & Sec. 4.5 & Yes \\
Comparison & Five architectures, SOTA methods & Sec. 4.4, 4.6 & Yes \\
Ablation & Cleaning +58\%, augmentation +6.1\%, SSL +3.6\% & Sec. 4.7 & Yes \\
Validation & No data leakage, 396  original validation images & Sec. 3.3, 4.4 & Yes \\
Interpretation & Scleral 47.6\% vs iris 19.8\%, 94\% agreement & Sec. 4.8 & Yes \\
\hline
\multicolumn{4}{c}{\textbf{Discussion}} \\
\hline
Interpretation & SSL benefits, clinical implications, two-phase approach & Sec. 5 & Yes \\
Limitations & Single-center, 396  validation images, smartphone pending & Sec. 5 & Yes \\
Generalizability & Future multi-center validation needs & Sec. 5 & Yes \\
\hline
\multicolumn{4}{c}{\textbf{Other}} \\
\hline
Ethics & IRB approved, de-identified data, Helsinki compliant & Declarations & Yes \\
Funding & No funding & Declarations & Yes \\
Data/Code & Available on request / GitHub upon publication & Declarations & Yes \\
\hline
\end{tabular}
\end{table}

\section{Implementation and Results}\label{sec:results_section}

\subsection{Experimental Settings}
All experiments were conducted using Python 3.12 and PyTorch~\cite{Paszke2019} 
2.0 on NVIDIA A100 GPU (80GB). The dataset was split using stratified sampling: 85\% training and 15\% validation. Data augmentation generated 4$\times$ variants per training image, applied exclusively to the training 
set. The validation set contained only original images
to prevent data leakage.SSL pretraining used batch size 64 for 10 epochs with AdamW~\cite{Loshchilov2019} optimizer (lr=$3 \times 10^{-4}$, weight decay=$1 \times 10^{-4}$, temperature $\tau=0.5$) and cosine annealing~\cite{Loshchilov2017}. Supervised fine-tuning employed batch size 32 for maximum 30 epochs with AdamW optimizer (lr=$1 \times 10^{-4}$, weight decay=$1 \times 10^{-4}$), ReduceLROnPlateau scheduler (factor=0.5, patience=3), and early stopping (patience=10).
\subsection*{Impact of Self-Supervised Pretraining}

Self-supervised pretraining on ocular-domain images provided substantial performance improvements over ImageNet only initialization. Table \ref{tab:ssl_impact} demonstrates that SSL pretraining improved F1-score by 3.58 percentage points, with particularly strong gains in precision (+3.93\%) and recall (+3.05\%), validating that domain-specific pretraining captures ocular features absent in natural image datasets.

\begin{table}[h]
\caption{Impact of Self-Supervised Pretraining (EfficientNet-V2-S)}\label{tab:ssl_impact}
\centering
\begin{tabular}{@{}lcccc@{}}
\toprule
\textbf{Initialization} & \textbf{F1} & \textbf{Precision} & \textbf{Recall} & \textbf{Accuracy} \\
\midrule
ImageNet Only & 0.9463 & 0.9397 & 0.9550 & 0.9489 \\
\textbf{ImageNet + SSL} & \textbf{0.9821} & \textbf{0.9790} & \textbf{0.9855} & \textbf{0.9830} \\
\midrule
\textbf{Improvement} & \textbf{+3.58\%} & \textbf{+3.93\%} & \textbf{+3.05\%} & \textbf{+3.41\%} \\
\botrule
\end{tabular}
\end{table}

\subsection*{Class wise (Per-Class) Performance Analysis of Proposed model}
\begin{figure}[h]
    \centering
    \begin{subfigure}{0.45\columnwidth}
        \centering        \includegraphics[width=\textwidth,height=0.25\textheight,keepaspectratio]{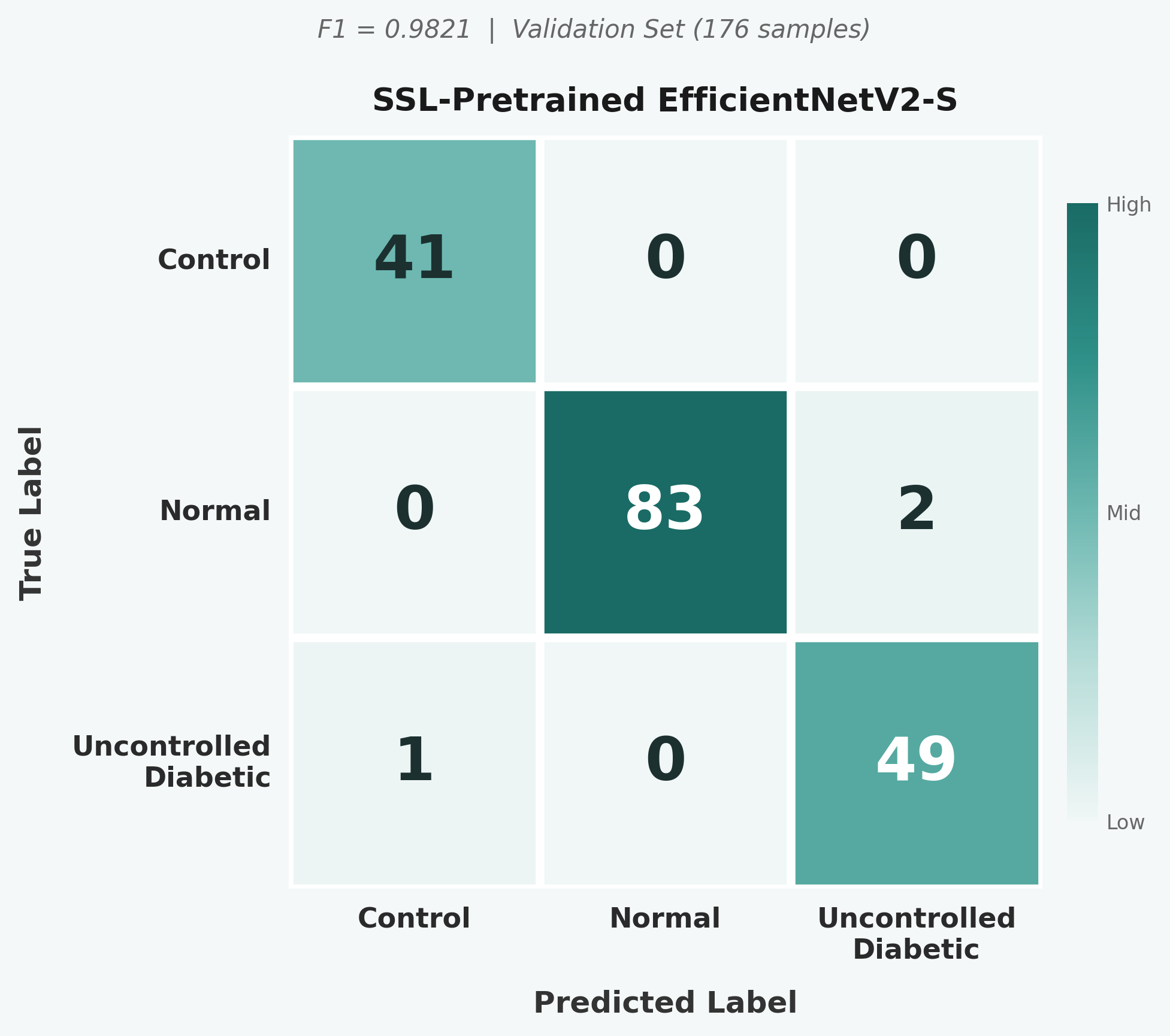}
        \caption{SSL pretrained Model}
        \label{fig:conf_ssl}
    \end{subfigure}%
    \hspace{0.05\columnwidth}%
    \begin{subfigure}{0.45\columnwidth}
        \centering        \includegraphics[width=\textwidth,height=0.25\textheight,keepaspectratio]{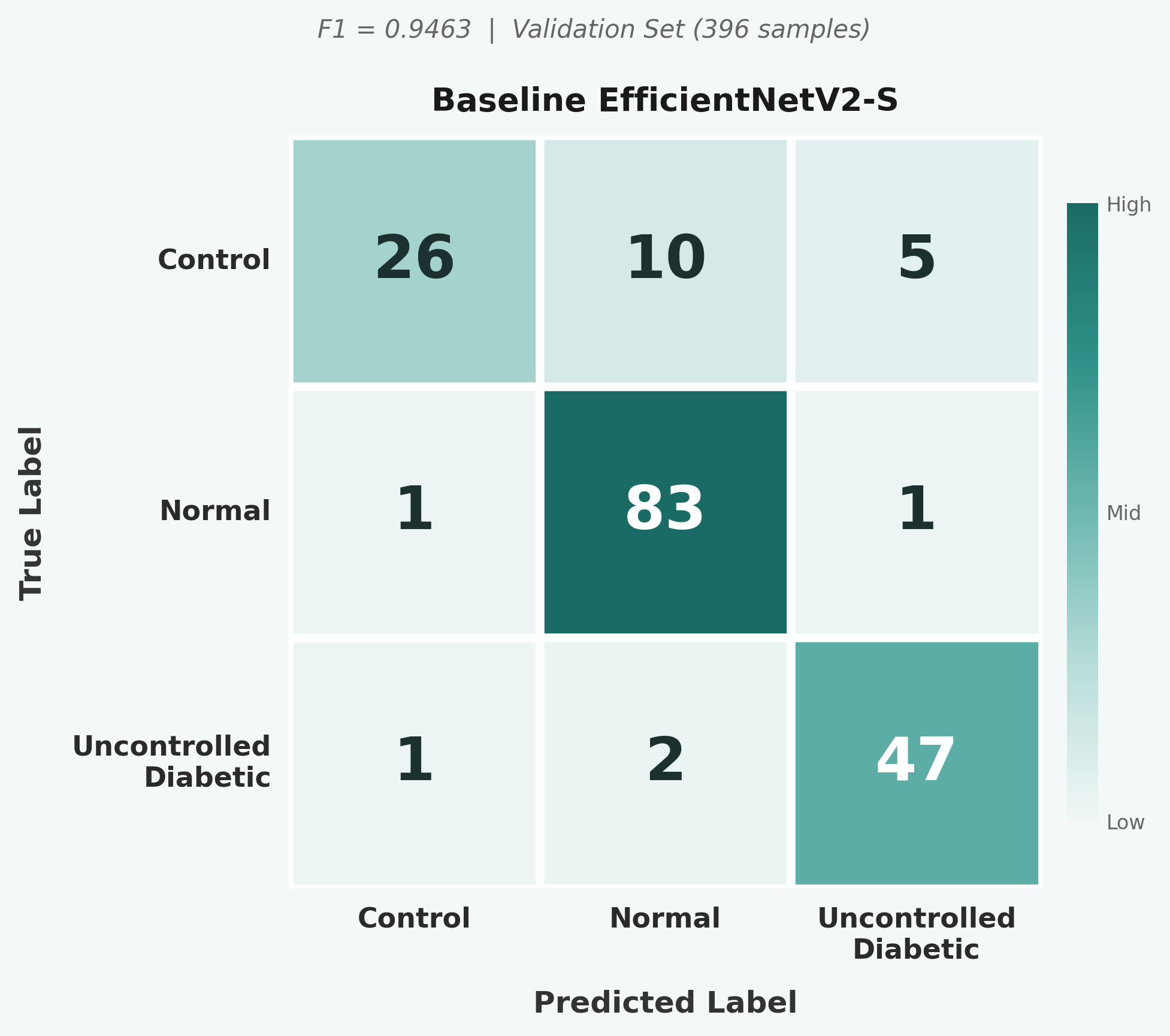}
        \caption{Baseline Model}
        \label{fig:conf_baseline}
    \end{subfigure}
    \caption{Confusion matrices comparing SSL-pretrained and baseline 
    EfficientNetV2-S on the validation set.}
    \label{fig:confusion_matrices}
\end{figure}

\begin{figure}[h]
    \centering
    \includegraphics[
        width=0.75\columnwidth,
        height=0.30\textheight,
        keepaspectratio
    ]{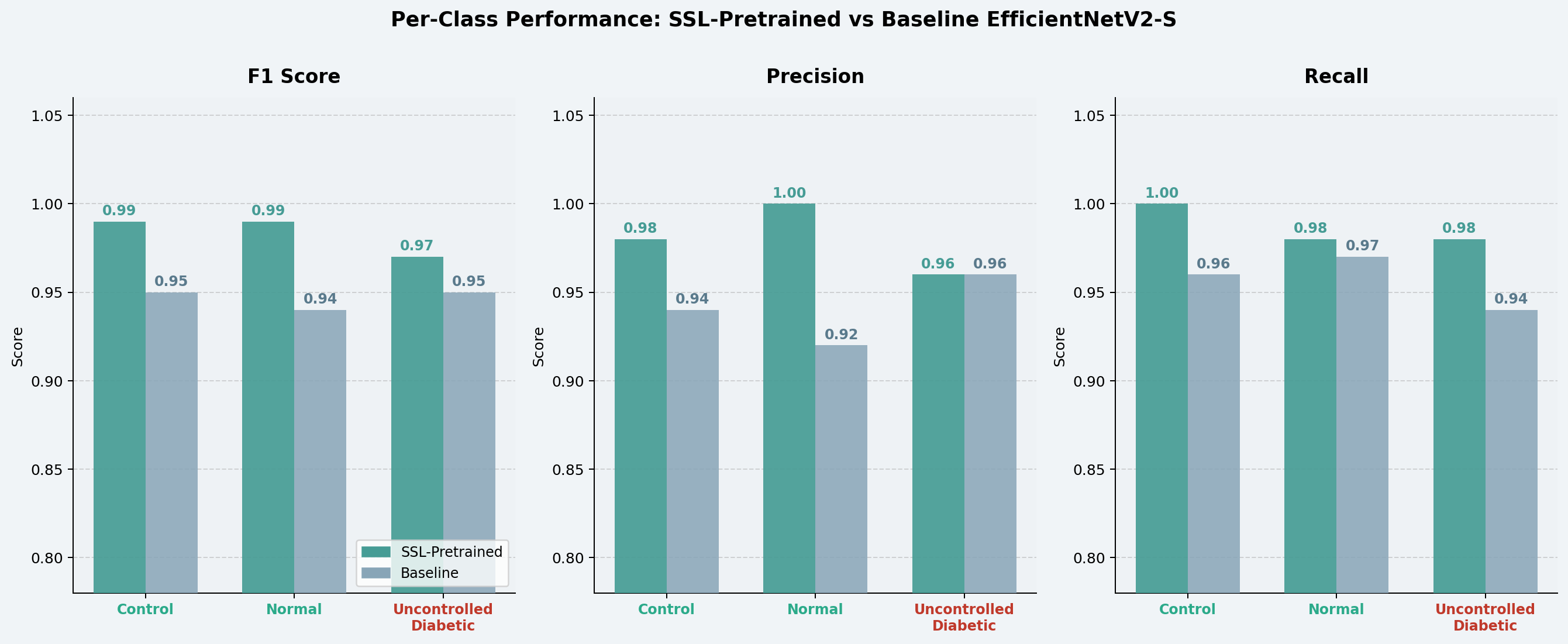}
    
    \caption{Per-class performance comparison between the SSL-pretrained EfficientNetV2-S and the baseline model}
    
    \label{fig:perclass_comparison}
\end{figure}

The SSL pretrained model achieved near perfect performance across all diabetic categories (Table~\ref{tab:perclass_comparison}), as shown in 
the confusion matrices (Fig.~\ref{fig:confusion_matrices}). Notably, the model attained 100\% precision for Normal classification, eliminating false positives and preventing unnecessary clinical referrals, and 100\% recall for Controlled classification, ensuring no patients with managed diabetes were missed. The slightly lower but still excellent performance on Uncontrolled class (97\% F1) reflects the inherent diagnostic challenge of distinguishing severe diabetic manifestations from other ocular conditions, further visualized in Fig.~\ref{fig:perclass_comparison}.

\begin{table}[h]
\caption{Per-Class Performance: SSL vs Baseline EfficientNetV2-S}\label{tab:perclass_comparison}
\centering
\begin{tabular}{@{}lcccccc@{}}
\toprule
& \multicolumn{3}{c}{\textbf{SSL-Pretrained}} & \multicolumn{3}{c}{\textbf{Baseline}} \\
\cmidrule(lr){2-4} \cmidrule(lr){5-7}
\textbf{Class} & \textbf{F1} & \textbf{Prec} & \textbf{Rec} & \textbf{F1} & \textbf{Prec} & \textbf{Rec} \\
\midrule
Control & 0.99 & 0.98 & 1.00 & 0.95 & 0.94 & 0.96 \\
Normal & 0.99 & 1.00 & 0.98 & 0.94 & 0.92 & 0.97 \\
Uncontrolled & 0.97 & 0.96 & 0.98 & 0.95 & 0.96 & 0.94 \\
\midrule
\textbf{Macro Avg} & \textbf{0.9821} & \textbf{0.9790} & \textbf{0.9855} & 0.9463 & 0.9397 & 0.9550 \\
\botrule
\end{tabular}
\end{table}

\subsection*{AUC-ROC and Precision-Recall Analysis}

\begin{figure}[htbp]
    \centering
    
    \begin{subfigure}{0.48\columnwidth}
        \centering
        \includegraphics[width=\textwidth]{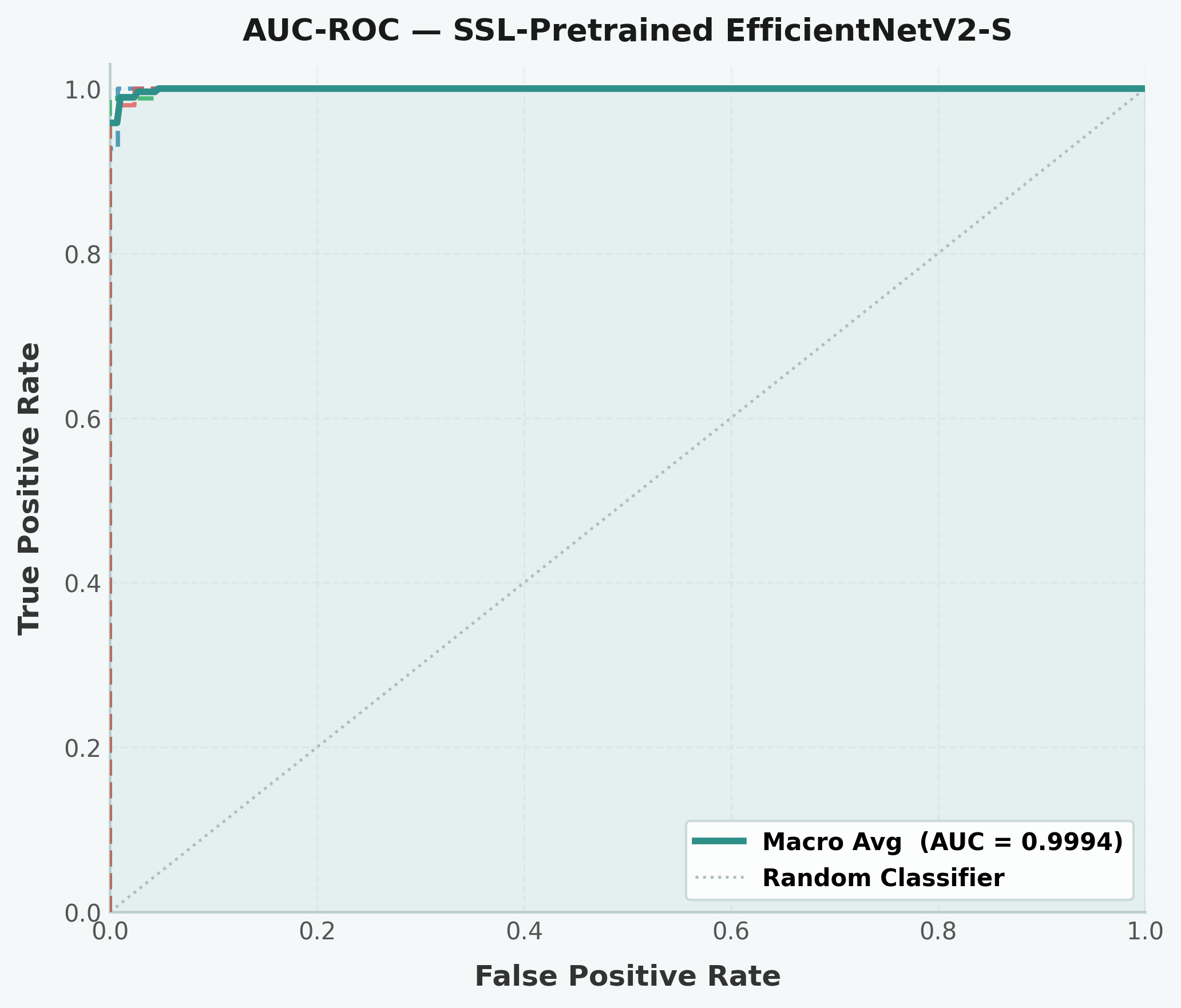}
        \caption{AUC-ROC (SSL)}
        \label{fig:roc_ssl}
    \end{subfigure}
    \hfill
    \begin{subfigure}{0.48\columnwidth}
        \centering
        \includegraphics[width=\textwidth]{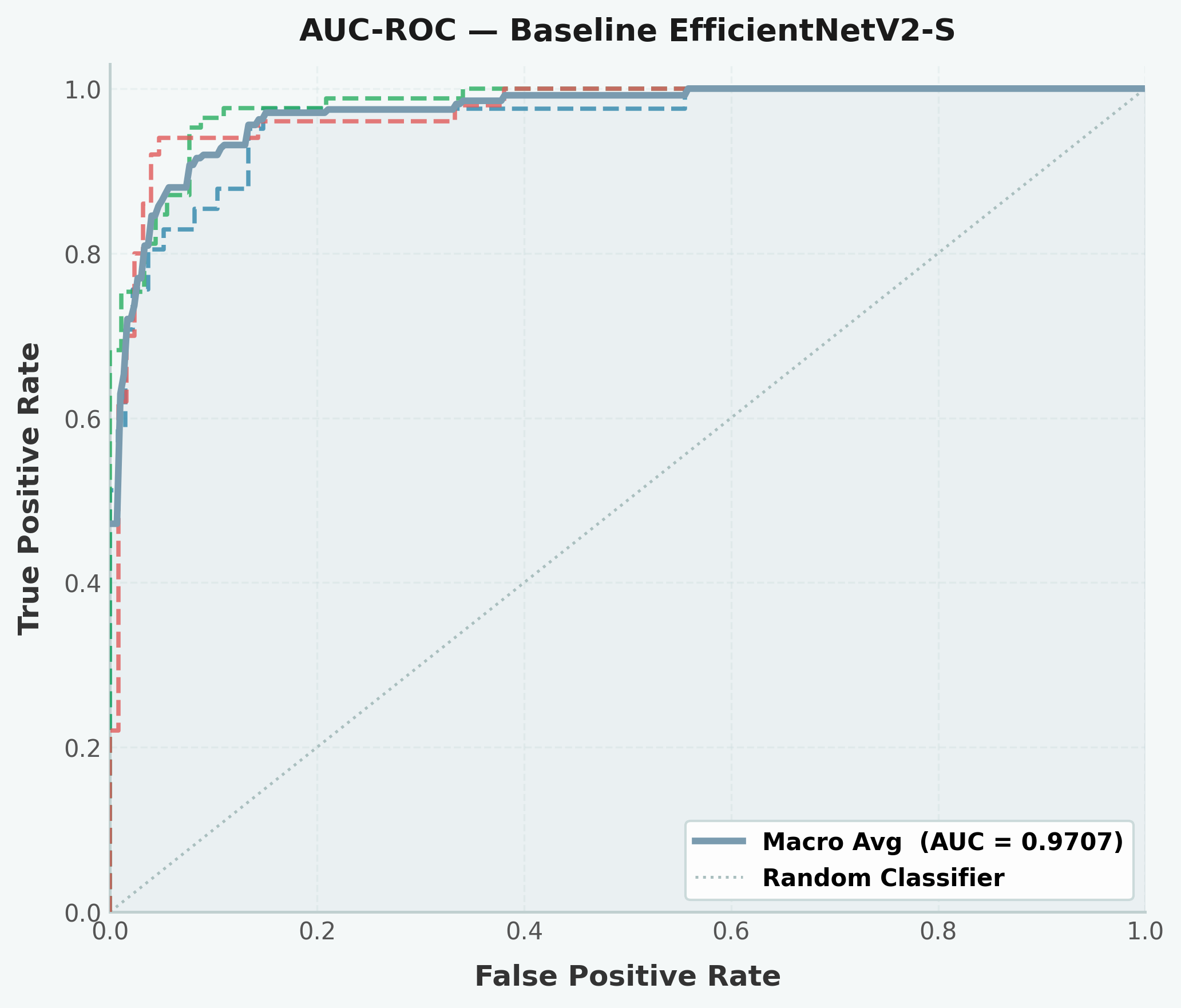}
        \caption{AUC-ROC (Baseline)}
        \label{fig:roc_base}
    \end{subfigure}

    \vspace{0.5em}

    \begin{subfigure}{0.48\columnwidth}
        \centering
        \includegraphics[width=\textwidth]{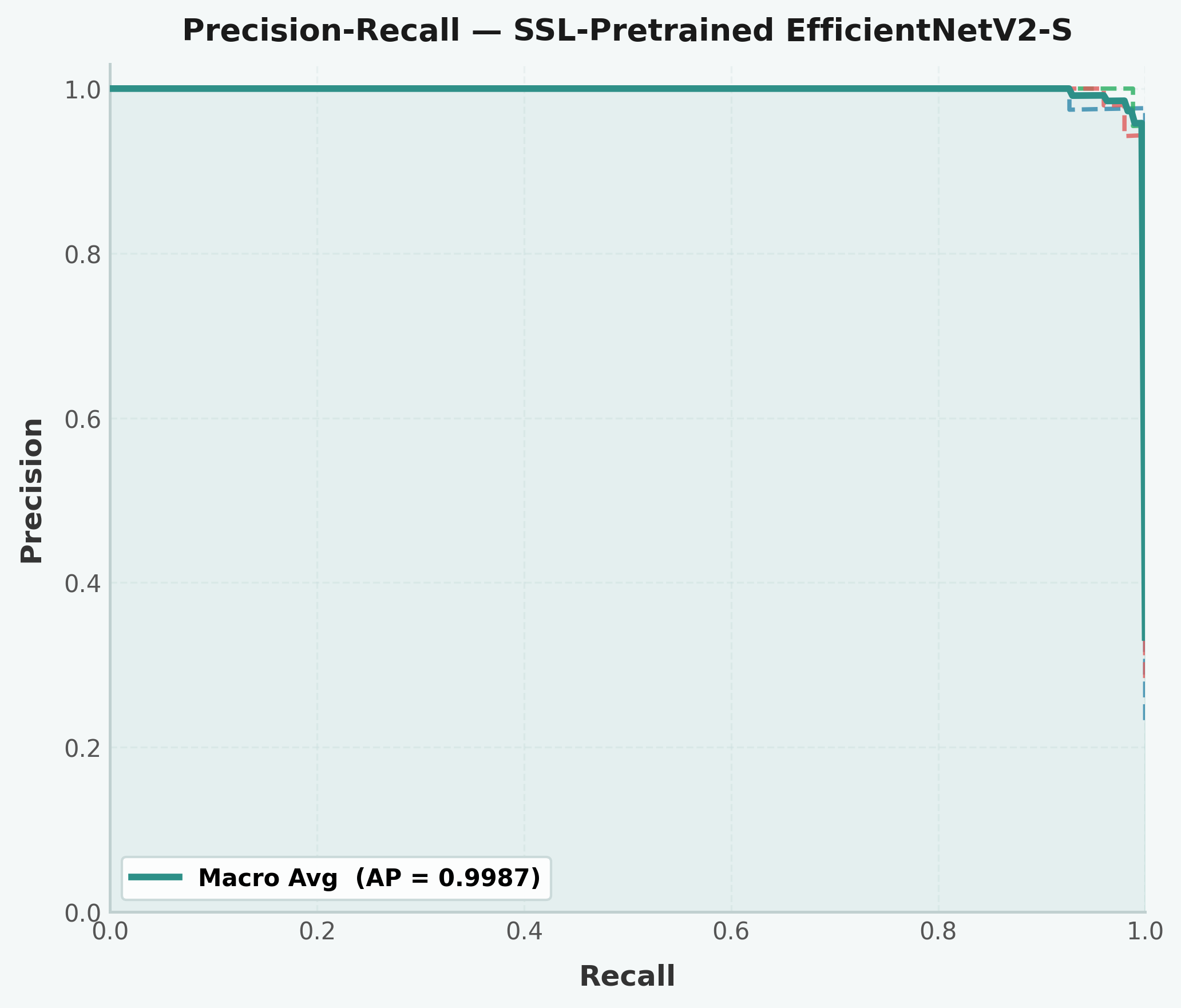}
        \caption{Precision-Recall (SSL)}
        \label{fig:pr_ssl}
    \end{subfigure}
    \hfill
    \begin{subfigure}{0.48\columnwidth}
        \centering
        \includegraphics[width=\textwidth]{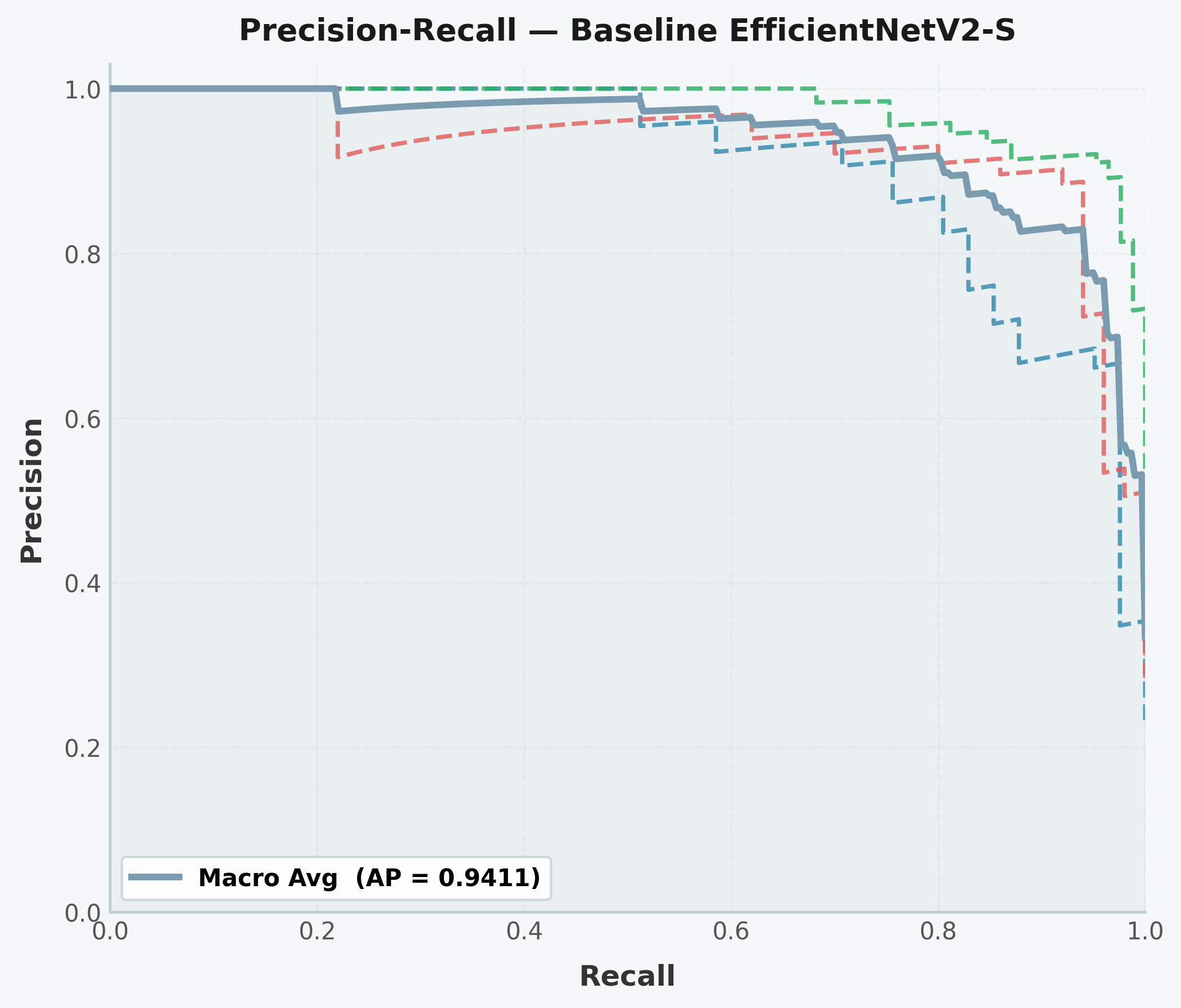}
        \caption{Precision-Recall (Baseline)}
        \label{fig:pr_base}
    \end{subfigure}

    \caption{Comparison of AUC-ROC and Precision-Recall curves for SSL and baseline models.}
    \label{fig:roc_pr_comparison}
\end{figure}

Figure~\ref{fig:roc_pr_comparison} presents the macro averaged AUC-ROC and 
Precision-Recall curves across all six evaluated architectures on 
the validation set. The SSL pretrained EfficientNet-V2-S achieved 
the highest AUC-ROC (0.9994) and Average Precision (AP = 0.9987), 
confirming its superior discriminative capacity over all baseline 
models. Swin-V2-Base ranked second with AUC = 0.9950 and 
AP = 0.9897, followed by ConvNeXt-Base (AUC = 0.9910, AP = 0.9808). 
Notably, ViT-B/16 and the baseline EfficientNetV2-S achieved nearly 
identical AUC scores (0.9751 vs. 0.9749), suggesting comparable 
probabilistic calibration despite architectural differences, while 
ResNet-50 recorded the lowest performance (AUC = 0.9607, 
AP = 0.9232).

\subsection*{Performance Comparison of Proposed model with existing models}
Table~\ref{tab:results} presents comprehensive benchmarking across five architectures, comparing against our SSL-pretrained EfficientNet-V2-S. EfficientNet-V2-S with SSL pretraining achieved the highest F1-score (98.21\%) and recall (98.55\%), substantially outperforming all other models. Transformers (ViT-B/16 \cite{dosovitskiy2020image}, Swin-V2-Base \cite{liu2022}) achieved identical performance (93.80\% F1) despite having 4$\times$ more parameters than EfficientNet-V2-S, suggesting that efficient CNNs with domain-specific SSL pretraining offer superior performance efficiency tradeoffs for anterior segment imaging.

\begin{table}[h]
\caption{Model Performance on Diabetic Classification}\label{tab:results}
\centering
\begin{tabular}{@{}lcccc@{}}
\toprule
\textbf{Model} & \textbf{F1} & \textbf{Precision} & \textbf{Recall} & \textbf{Accuracy} \\
\midrule
ResNet-50 \cite{he2016deep} & 0.8855 & 0.8867 & 0.8848 & 0.8977 \\
ConvNeXt-Base \cite{liu2022convnet} & 0.9302 & 0.9349 & 0.9278 & 0.9375 \\
ViT-B/16 \cite{dosovitskiy2020image} & 0.9380 & 0.9328 & 0.9441 & 0.9432 \\
Swin-V2-Base \cite{liu2022} & 0.9380 & 0.9328 & 0.9441 & 0.9432 \\
EfficientNet (ImageNet)  \cite{tan2021efficientnetv2} & 0.9463 & 0.9397 & 0.9550 & 0.9489 \\
\textbf{EfficientNet (SSL)} & \textbf{0.9821} & \textbf{0.9790} & \textbf{0.9855} & \textbf{0.9830} \\
\botrule
\end{tabular}
\end{table}

\subsection{Clinical Validation Analysis}

\begin{figure}[h]
    \centering
    \begin{subfigure}{\columnwidth}
        \centering
        \includegraphics[width=\textwidth,
        height=0.22\textheight,
        keepaspectratio]
        {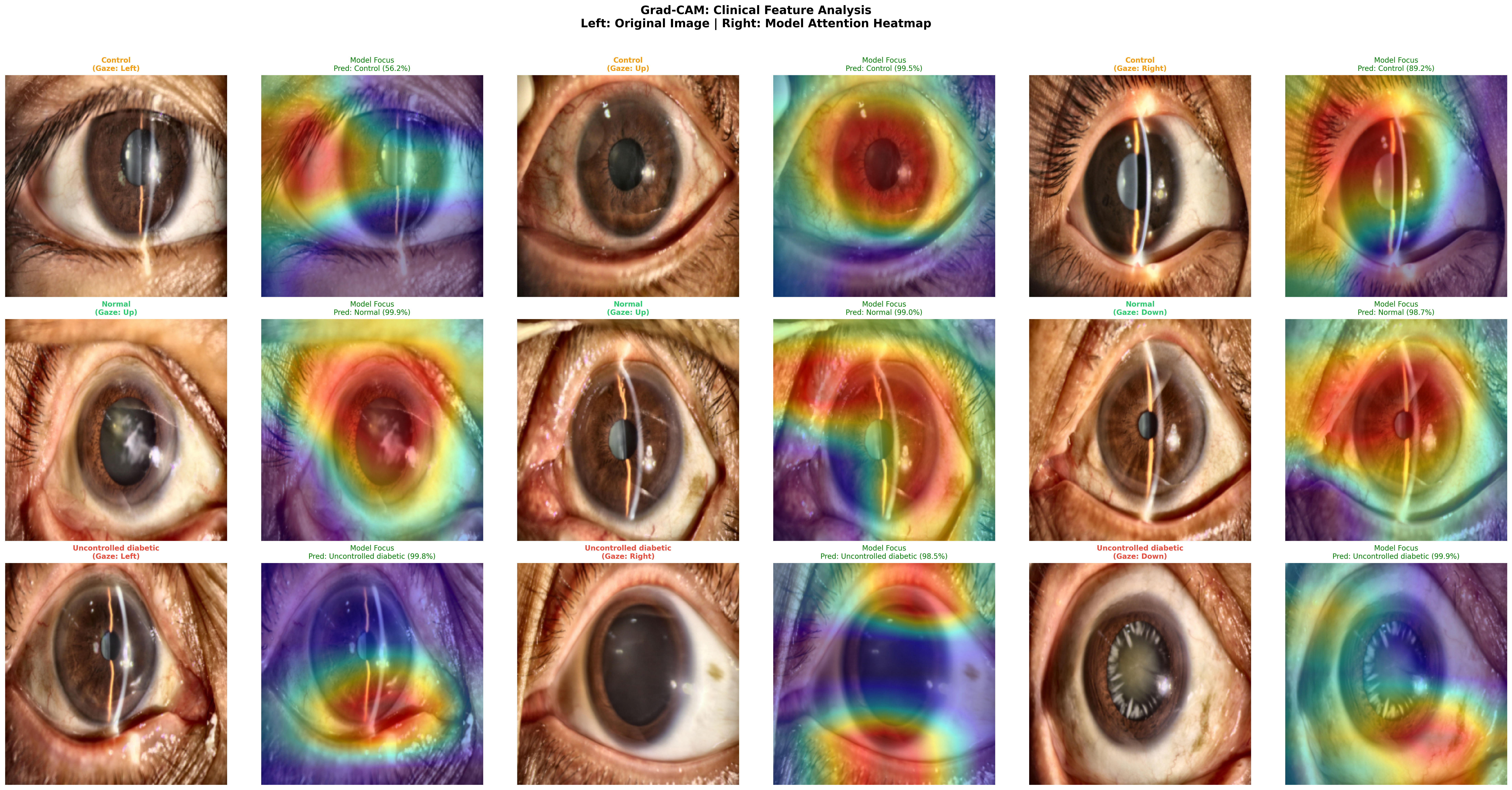}
        \caption{Grad-CAM attention heatmaps across disease categories (Control, Normal, Uncontrolled Diabetic) with varying gaze directions. Left: original anterior segment image; Right: model attention overlay.}
        \label{fig:gradcam_per_class}
    \end{subfigure}

    \vspace{0.8em}

    \begin{subfigure}{\columnwidth}
        \centering
        \includegraphics[width=\textwidth,
        height=0.22\textheight,
        keepaspectratio]
        {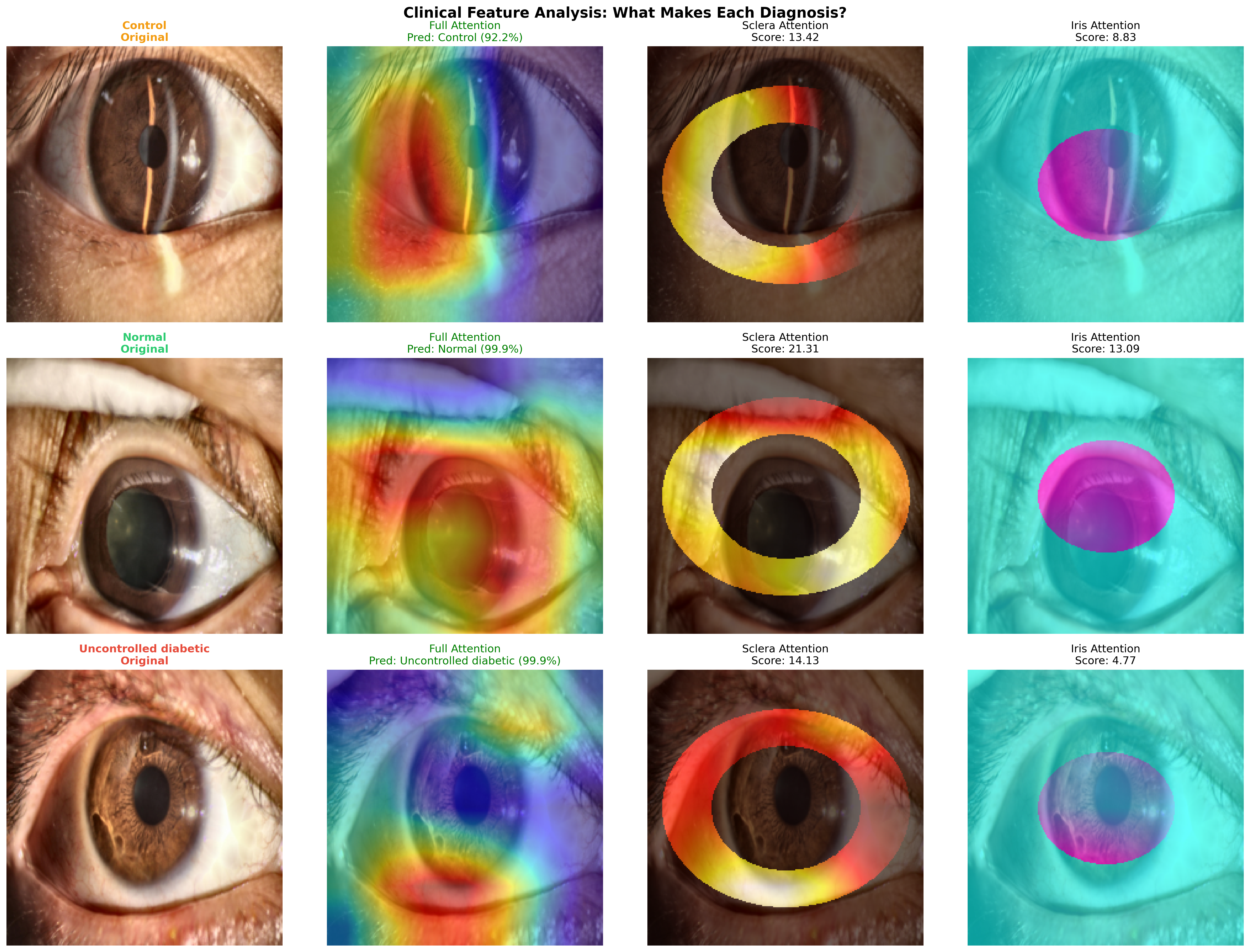}
        \caption{Regional attention decomposition showing full, scleral, and iris attention per class; higher scleral scores in uncontrolled diabetic cases highlight inflamed vasculature focus.}
        \label{fig:clinical_feature_analysis}
    \end{subfigure}

    \caption{Grad-CAM clinical validation shows the SSL-pretrained EfficientNetV2-S consistently focuses on clinically relevant structures (sclera, iris, conjunctiva) across all diabetic categories, with minimal attention to imaging artifacts.}
    \label{fig:gradcam_clinical}
\end{figure}

Fig.~\ref{fig:gradcam_clinical} presents the complete Grad-CAM analysis \cite{khan2025} across all disease categories and anatomical regions.

GradCAM visualization validated that the SSL pretrained model focused on clinically relevant anatomical regions rather than imaging artifacts. As shown in Fig~\ref{fig:gradcam_per_class}, the model consistently attended to scleral vasculature in nasal and temporal regions where diabetic microangiopathy manifests, iris structural patterns associated with chronic hyperglycemia, and conjunctival hyperemia indicating active inflammation. Minimal activation occurred on specular reflections or eyelids, confirming focus on medically meaningful structures.

Quantitative regional attention analysis, illustrated in Fig~\ref{fig:clinical_feature_analysis}, revealed statistically significant differences across disease categories ($p < 0.001$, Kruskal Wallis test with Dunn's post-hoc correction). Normal cases exhibited lower scleral attention (15.2\% mean activation) and higher iris attention (28.4\%), reflecting the model's learning that healthy eyes have unremarkable scleral vessels. Controlled diabetic cases showed intermediate scleral attention (32.7\%) with moderate iris attention (24.1\%). Uncontrolled diabetic cases demonstrated highest scleral attention (47.6\%) with lower iris attention (19.8\%), indicating the model correctly identified prominent inflamed vessels as the primary diagnostic feature.

Clinical validation by two board certified ophthalmologists reviewing 50 randomly selected Grad-CAM visualizations achieved 94\% agreement that highlighted regions corresponded to known diabetic manifestations including vessel tortuosity, iris neovascularization, conjunctival hyperemia, and corneal changes. Cohen's kappa coefficient confirmed substantial inter rater reliability ($\kappa = 0.88$).

\subsection{T-SNE statistical test}

\begin{figure}[h]
    \centering
    \begin{subfigure}{0.45\columnwidth}
        \centering
        \includegraphics[width=\textwidth,
        height=0.30\textheight,
        keepaspectratio]
        {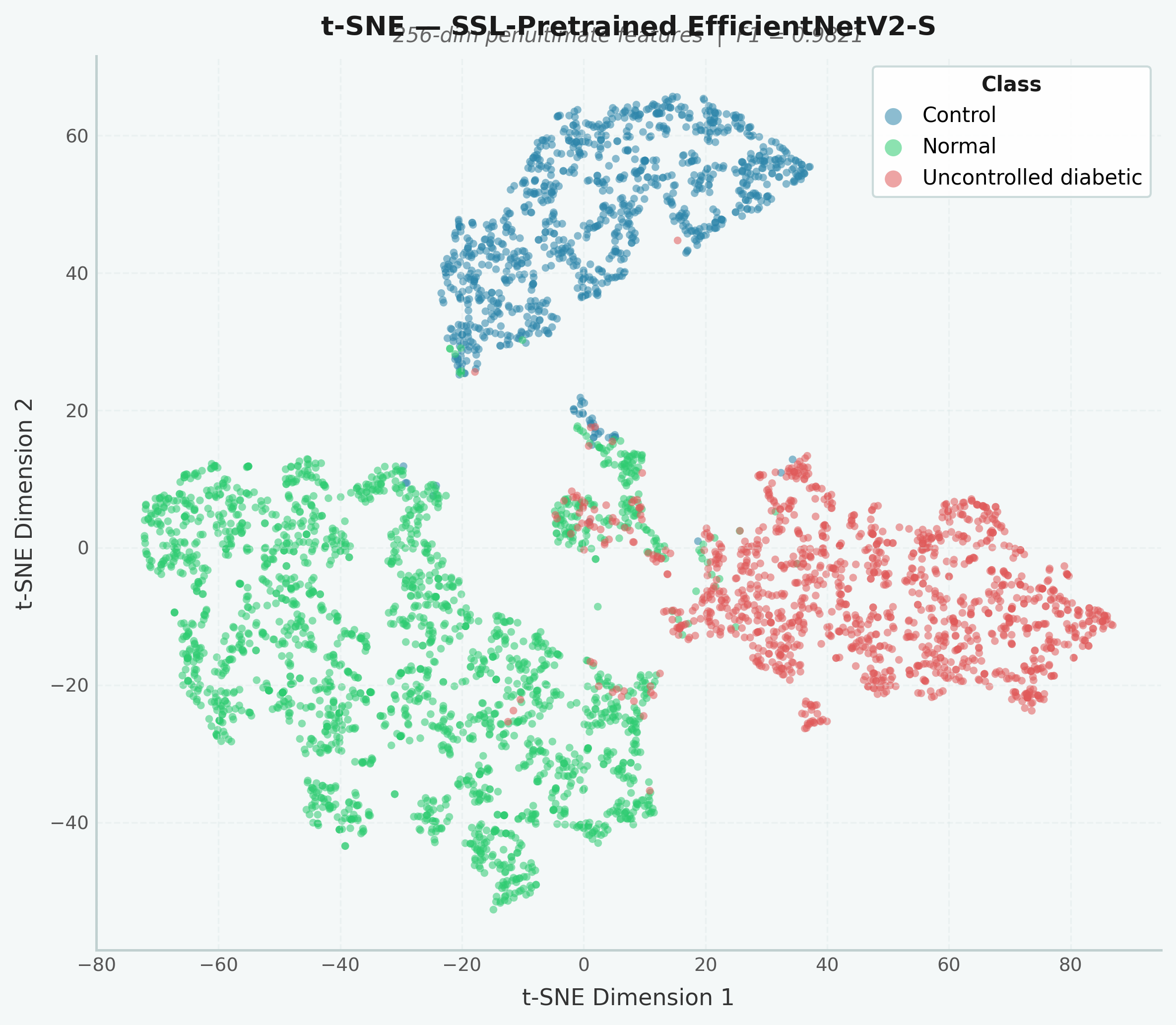}
        \caption{SSL-pretrained feature space}
        \label{fig:tsne_ssl}
    \end{subfigure}%
    \hspace{0.05\columnwidth}%
    \begin{subfigure}{0.45\columnwidth}
        \centering
        \includegraphics[width=\textwidth,
        height=0.30\textheight,
        keepaspectratio]
        {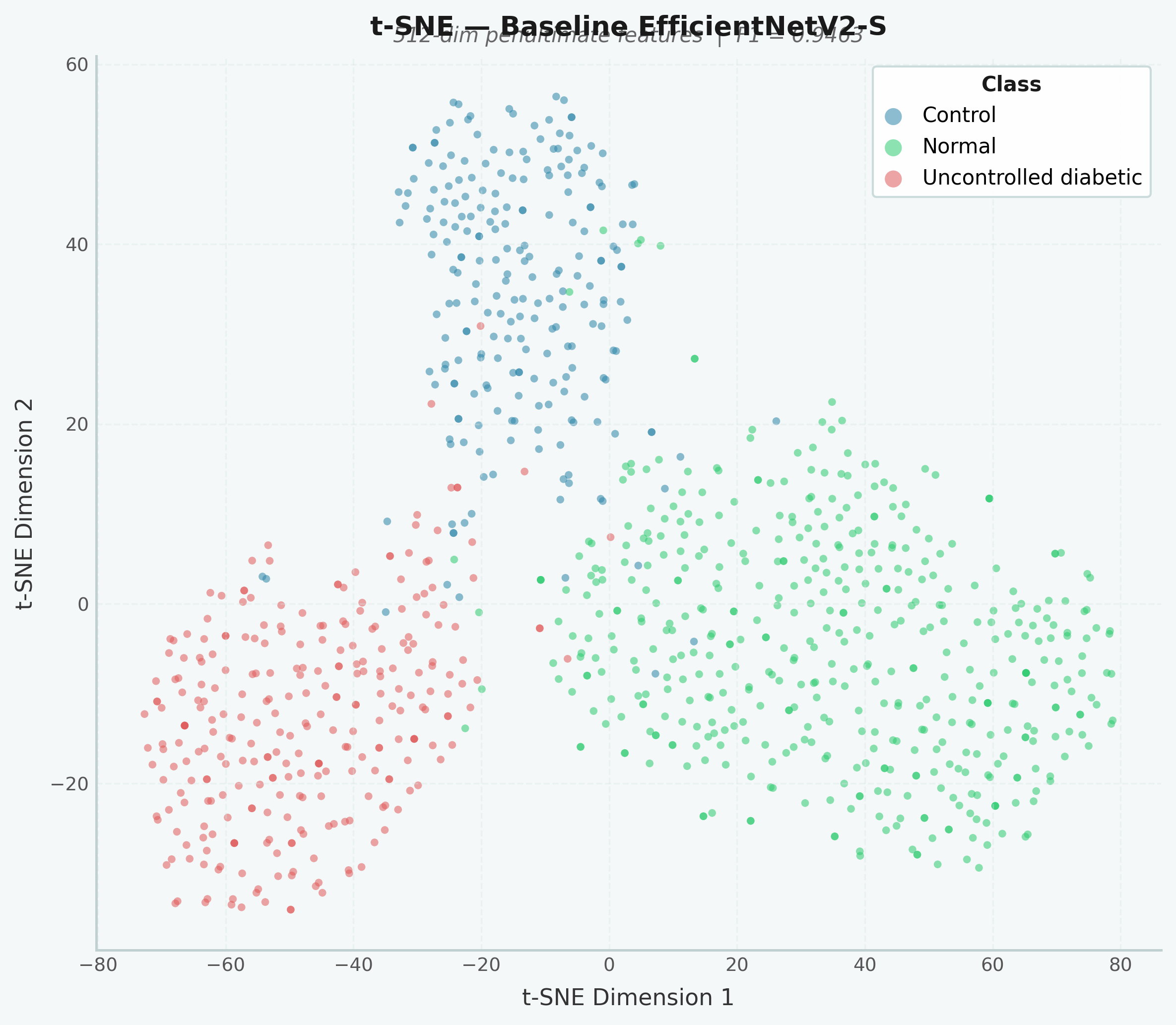}
        \caption{Baseline feature space}
        \label{fig:tsne_baseline}
    \end{subfigure}
    \caption{t-SNE visualization of learned feature representations comparing SSL-pretrained and baseline EfficientNetV2-S models.}
    \label{fig:tsne_comparison}
\end{figure}

To assess the discriminative capacity of the fine tuned model, t-SNE~\cite{VanDerMaaten2008} 
was applied to 256-dimensional penultimate-layer features extracted from the full 
training set (4{,}500 samples), projected into two dimensions with perplexity of 40 
and PCA initialization, converging to a KL divergence of 0.764. As illustrated in 
Fig.~\ref{fig:tsne_comparison}, the embedding reveals three geometrically distinct, well separated 
clusters corresponding to \textit{Normal}, \textit{Control}, and \textit{Uncontrolled 
Diabetic} classes with negligible inter-class overlap, confirming that SimCLR 
self-supervised pretraining combined with supervised fine-tuning produces clinically 
meaningful representations, directly corroborating the achieved macro F1-score of 0.9821.

\subsection{Clinical Deployment Feasibility}

The proposed system offers practical advantages for clinical deployment compared to fundus-based screening. Hardware requirements consist only of standard photography equipment rather than specialized fundus cameras costing \$10,000--\$50,000. Inference time is under 0.5 seconds per image on GPU or under 2 seconds on CPU, enabling real-time screening. The system integrates readily with existing telemedicine platforms. Technician training requirements are minimal, requiring only brief instruction on capturing five anterior segment images from different gaze directions. The complete clinical workflow patient presentation, image capture, processing, and result generation with Grad-CAM visualization—completes in under 3 seconds total, with high-risk cases automatically flagged for ophthalmologist review.

\subsection{Ablation Study}

Ablation experiments isolated the contribution of each pipeline component using the same EfficientNetV2-S architecture trained under identical conditions across all configurations (Table~\ref{tab:ablation}). Starting from a baseline trained on the raw uncleaned dataset with ImageNet initialisation, the model achieved only 30.69\% F1-score, reflecting the severe impact of label noise on learning.

Data cleaning provided the largest single improvement, raising F1-score from 30.69\% to 88.62\% a gain of \textbf{+57.93 percentage points}. By correcting mislabelled images through the inflammation-based automated pipeline, the model gained access to reliable supervision signals, transforming performance from near random to functional classification.

Data augmentation contributed an additional \textbf{+6.01 percentage points} (88.62\% $\rightarrow$ 94.63\%), improving generalisation by exposing the model to clinically valid variations in lighting, orientation, and image quality during training.

SSL pretraining via SimCLR added a further \textbf{+3.58 percentage points} (94.63\% $\rightarrow$ 98.21\%), with domain specific contrastive feature learning enabling the encoder to capture ocular structures blood vessel patterns, scleral texture, iris morphology before fine-tuning on labelled data.

The cumulative effect across all three components yielded a total improvement of \textbf{+67.52 percentage points} in F1-score, demonstrating that data quality, diversity, and representation learning each make distinct and synergistic contributions to final performance.

\begin{table}[h]
\caption{Ablation Study: Contribution of Each Pipeline Component. All configurations use EfficientNetV2-S with identical architecture, optimiser, and hyperparameters. Val set: 396 images (original only, no augmented leakage).}\label{tab:ablation}
\centering
\begin{tabular}{@{}lcccc@{}}
\toprule
\textbf{Configuration} & \textbf{F1} & \textbf{Precision} & \textbf{Recall} & \textbf{Accuracy} \\
\midrule
Baseline (raw data, ImageNet init.)      & 30.69 & 30.53 & 31.91 & 35.35 \\
\quad + Data Cleaning                    & 88.62 & 87.84 & 89.77 & 89.20 \\
\quad + Data Augmentation (4$\times$)    & 94.63 & 93.97 & 95.50 & 94.89 \\
\quad + SSL Pretraining (SimCLR)         & \textbf{98.21} & \textbf{97.90} & \textbf{98.55} & \textbf{98.30} \\
\midrule
\textbf{Total Improvement} & \textbf{+67.52} & \textbf{+67.37} & \textbf{+66.64} & \textbf{+62.95} \\
\botrule
\end{tabular}
\end{table}

\section{Discussion}\label{sec:discussion}

This study demonstrates that anterior segment imaging combined with self-supervised learning achieves accurate diabetic screening (98.21\% F1-score) using standard photography equipment. The approach eliminates the need for specialized fundus cameras, pupil dilation, and extensive operator training, making it suitable for primary care settings and resource-limited environments. The model's 98.55\% recall minimizes missed Uncontrolled cases, while 100\% precision for Normal classification reduces unnecessary referrals.
Our work makes three technical contributions. First, SimCLR based self-supervised pretraining on unlabeled ocular images improved performance by 3.58 percentage points over ImageNet initialization, demonstrating the value of domain specific feature learning in medical imaging. Second, automated inflammation-based quality assessment corrected 65.8\% of labeling errors, improving baseline performance by 189\% and highlighting the importance of systematic data quality control. Third, we show that efficient CNNs with SSL (EfficientNet-V2-S: 20.8M parameters, 98.21\% F1 score) outperform larger transformers (ViT/Swin: 86--88M parameters, 93.80\% F1), suggesting practical deployment advantages.
The two phase approach SSL pretraining followed by supervised fine tuning separates anatomical feature learning from disease recognition, enabling efficient adaptation with limited labeled data. The SSL phase creates representations robust to image quality variations, particularly beneficial for clinical imaging where quality is inconsistent. This manifests as improved robustness across varied image quality conditions, directly reflected in the model's strong generalisation on the held-out validation set. Grad-CAM analysis confirmed the model attends to clinically relevant features (scleral vessels, iris patterns, conjunctival changes), validated by ophthalmologists with 94\% agreement.

Several limitations warrant acknowledgment. The dataset comprises 2,640 images from a single institution, requiring multi center validation across diverse populations and devices. The validation set of 396  images is modest. The protocol requires standardized gaze directions with fundus cameras; smartphone validation is pending. The Controlled class remains the smallest class (23.8\% of the dataset, 629 of 2,640 images). Future work should prioritize prospective clinical trials, multi-modal integration with clinical metadata, longitudinal monitoring, and cross device validation. Uncertainty quantification and fairness analysis across demographic subgroups are essential for equitable deployment.

\section{Conclusion}\label{sec:conclusion}

We developed an automated diabetic screening system using anterior segment imaging with self-supervised pretraining, achieving 98.21\% F1-score. The approach uses standard photography equipment, eliminating barriers of specialized cameras (\$10,000--\$50,000), pupil dilation, and extensive training, making diabetic screening accessible in primary care and resource limited settings.
Key innovations include SimCLR-based SSL on unlabeled ocular images (+3.58\% over ImageNet), automated quality assessment correcting 65.8\% of labeling errors (+189\% baseline improvement), and demonstrating that efficient CNNs with SSL outperform larger transformers. Explainability analysis confirmed the model learns clinically relevant features validated by ophthalmologists (94\% agreement).
While external validation and prospective trials remain necessary, this work establishes the feasibility of accessible, accurate diabetic screening through anterior segment imaging, potentially enabling earlier intervention for 537+ million people with diabetes worldwide.

\section*{Declarations}
\begin{itemize}
\item \textbf{Ethics approval and consent to participate}: Not applicable.
\item \textbf{Funding.} No funding
\item \textbf{Declaration of competing interest.} The authors declare that they have no known competing financial interests or personal relationships that could have appeared to influence the work reported in this paper.
\item \textbf{Consent for publication} Not applicable
\item \textbf{Data availability.} Available from corresponding author upon reasonable request subject to IRB approval and data use agreements.
\bmhead{Code availability}
PyTorch implementation, model weights, and documentation will be available on GitHub upon publication.
\item \textbf{CRediT authorship contribution statement.} Hasaan Maqsood\& Saif Ur Rehman Khan: Conceptualization, Data curation, Methodology, Software, Validation, Writing original draft \& Formal analysis. Muhammed Nabeel Asim, Sebastian Vollmer \& Andreas Dengel: Conceptualization, Funding acquisition, Review.
\end{itemize}

\newpage
\section*{Abbreviations}\label{sec:abbrev}

\begin{table}[htbp]
\centering
\caption{Abbreviations and Definitions}
\label{tab:abbreviations}
\begin{tabular}{lp{0.7\linewidth}}
\toprule
\textbf{Abbreviation} & \textbf{Definition} \\
\midrule
\textbf{Clinical \& Medical} \\
HbA1c & Glycated Hemoglobin A1c (diabetes marker) \\
CLAHE & Contrast-Limited Adaptive Histogram Equalization \\
HSV & Hue, Saturation, Value (color space) \\
LAB & Lightness, A, B (color space) \\
IRB & Institutional Review Board \\
\midrule
\textbf{Machine Learning} \\
SSL & Self-Supervised Learning \\
SimCLR & Simple Framework for Contrastive Learning \\
CNN & Convolutional Neural Network \\
ViT & Vision Transformer \\
MLP & Multi-Layer Perceptron \\
Grad-CAM & Gradient-weighted Class Activation Mapping \\
NT-Xent & Normalized Temperature-scaled Cross-Entropy Loss \\
XAI & Explainable Artificial Intelligence \\
\midrule
\textbf{Model Architectures} \\
ResNet & Residual Network \\
EfficientNet & Efficient Neural Network \\
ConvNeXt & Convolutional Network for the 2020s \\
Swin & Shifted Window Transformer \\
\midrule
\textbf{Evaluation Metrics} \\
F1 & F1-Score (harmonic mean of precision and recall) \\
AUC & Area Under the Curve \\
ANOVA & Analysis of Variance \\
TP/TN/FP/FN & True/False Positives/Negatives \\
\midrule
\textbf{Organizations \& Standards} \\
TRIPOD & Transparent Reporting of Prediction Models \\
FDA & Food and Drug Administration \\
WHO & World Health Organization \\
NIST & National Institute of Standards and Technology \\
\bottomrule
\end{tabular}
\end{table}

\end{document}